\documentclass{article}
\usepackage{ijcai16}

\usepackage{times}

\usepackage{latexsym}
\usepackage{color}
\usepackage{graphicx}
\usepackage{amssymb,amsmath}
\usepackage[linesnumbered, ruled, vlined]{algorithm2e} 

\newcommand{\hide}[1]{} \newcommand{\vpara}[1]{\vspace{0.1in}\noindent\textbf{#1 }}
\newcommand{\beq}[1]{\begin{equation}#1\end{equation}}
\newcommand{\besp}[1]{\begin{split}#1\end{split}}

\title{Multi-Modal Bayesian Embeddings for Learning Social Knowledge Graphs}
\author{Zhilin Yang$^{\ddagger \dagger}$ ~~ Jie Tang$^\ddagger$ ~~ William Cohen$^\dagger$ \\ 
$^\ddagger$Tsinghua University  ~~ $^\dagger$Carnegie Mellon University\\
jietang@tsinghua.edu.cn ~ \{zhiliny,wcohen\}@cs.cmu.edu}

\begin{document}

\maketitle

\begin{abstract}
We study the extent to which online social networks can be connected to  knowledge bases. The problem is referred to as \textit{learning social knowledge graphs}.
We propose a multi-modal Bayesian embedding model, \textbf{GenVector}, to learn latent topics that generate word embeddings and network embeddings simultaneously.
GenVector leverages large-scale unlabeled data with embeddings and represents data of two modalities---i.e., social network users and knowledge concepts---in a shared latent topic space.
Experiments on three datasets show that the proposed method clearly outperforms state-of-the-art methods. We then deploy the method on AMiner, an online academic search system to connect with a network of 38,049,189 researchers with a knowledge base with 35,415,011 concepts. Our method significantly decreases the error rate of learning social knowledge graphs in an online A/B test with live users.

\hide{
Understanding the semantics of social networks is important for social mining. However, few works have studied how to bridge the gap between large-scale social networks and abundant collective knowledge, such as knowledge bases and encyclopedias. In this work, we study the problem of learning social knowledge graphs, which aims to accurately connect social network vertices to knowledge concepts.

We propose a multi-modal Bayesian embedding model, GenVector, to jointly incorporate topic models and word/network embeddings. The model leverages large-scale unlabeled data by incorporating embeddings, and co-represents data of two modalities, social network vertices and knowledge concepts, in a shared latent topic space.

Experiments on three datasets show that our method outperforms state-of-the-art methods. We deploy our algorithm on a large-scale academic social network by linking 39 million researchers to 35 million knowledge concepts, and decrease the error rate by 67\% according to an online test.
}
\end{abstract} 
\section{Introduction}

With the rapid development of online social networks, understanding user behaviors
and network dynamics
becomes an important yet challenging issue for social network mining.
Quite a few research works have been conducted towards dealing with this problem. For instance, 
Want et al.~\shortcite{wang2014mmrate} developed an approach to infer topic-based diffusion networks by considering different cascaded processes.
Han and Tang~\shortcite{HanTang:15KDD} proposed a probabilistic framework to model social links, communities, user attributes, roles and behaviors in a unified manner.
Sudhof et al.~\shortcite{sudhof2014sentiment} developed a theory of conditional dependencies between
human emotional states and implemented the theory using conditional random
fields (CRFs).
However, all the aforementioned works do not consider linking social contents to a universal knowledge bases, and thus the mining results can only be applied to a specific social network.
Tang et al.~\shortcite{Tang:15WWW} proposed the SOCINST model to extract entity information by incorporating both social context and domain knowledge.
However, users are not directly linked to knowledge bases, which limits deeper user understanding.

\hide{
fundamental
~\cite{wang2014mmrate,sudhof2014sentiment,nguyen2015topic}.

A large volume of information emerges and diffuses across large-scale social networks.

With the rapid development of online social networks, a large volume of information emerges and diffuses across large-scale social networks.

Mining knowledge from information and understanding social network users based on social text are fundamental problems of social network mining and text mining \cite{wang2014mmrate,sudhof2014sentiment,nguyen2015topic}.

In parallel, abundant collective knowledge in knowledge bases and encyclopedias, like Freebase \cite{bollacker2008freebase}, YAGO \cite{suchanek2007yago} and Wikipedia, are widely available. However, there is not extensive research on how to link social networks to collective knowledge, to provide semantic understanding about social information and users.
}

To bridge the gap between social networks and knowledge bases, we formalize a novel problem of learning social knowledge graphs. More specifically, 
given a social network, a knowledge base, and text posted by users on social networks, we aim to link each social network user to a given number of knowledge concepts. For example, in an academic social network, the problem can be defined as linking each researcher to a number of knowledge concepts in Wikipedia to reflect the research interests. Learning social knowledge graphs has potential applications in user modeling, recommendation, and knowledge-based search~\cite{sigurbjornsson2008flickr,kasneci2008naga,tang2008arnetminer}.

Multi-modal topic models, such as author-topic models \cite{rosen2004author} and Corr-LDA \cite{blei2003modeling}, can be extended to model the two modalities---i.e., social network users and knowledge concepts---in our problem. However, topic models are usually trained on text, and it is difficult to leverage information in knowledge bases and the structure of social networks. Recent advances in embeddings \cite{mikolov2013distributed,bordes2013translating,perozzi2014deepwalk} proposed to learn embeddings for words, knowledge concepts, and nodes in networks, which captures continuous semantics from unlabeled data. However, these embedding techniques do not model multi-modal correlation and thus cannot be directly applied to multi-modal settings.

We propose GenVector, a multi-modal Bayesian embedding model, to learn social knowledge graphs. GenVector uses latent discrete topic variables to generate continuous word embeddings and network-based user embeddings. 
The model 
combines the advantages of topic models and word embeddings, and is able to model multi-modal data and continuous semantics.
We present an effective learning algorithm to iteratively update the latent topics and the embeddings.

We collect three 
datasets for evaluation. Experiments show that GenVector clearly outperforms state-of-the-art methods. We also deploy GenVector into an online academic search system to connect a  network of 38,049,189 researchers with a knowledge base with 35,415,011 concepts.
We carefully design an online A/B test to compare the proposed model with the original algorithm of the system. 
 Results show that GenVector significantly decreases the error rate by 67\%.
Our main contributions are as follows:
\begin{itemize}
\item We formalize the problem of learning social knowledge graphs with the goal of connecting large-scale social networks with open knowledge bases.

\item We propose GenVector, a novel multi-modal Bayesian embedding model to model multi-modal embeddings with a shared latent topic space. 

\item We show that GenVector outperforms state-of-the-art methods in the task of learning social knowledge graphs on three datasets and significantly decreases the error rate in an online A/B test on a real system.
\end{itemize} 
\section{Problem Formulation}

The input of our problem includes a social network, a knowledge base, and text posted by users of the social network.
The social network is denoted as $\mathcal{G}^r = (\mathcal{V}^r, \mathcal{E}^r)$, where $\mathcal{V}^r$ is a set of  users and $\mathcal{E}^r$ is a set of edges between the users, either directed or undirected. The knowledge base
is denoted as $\mathcal{G}^k = (\mathcal{V}^k, \mathcal{C})$, where $\mathcal{V}^k$ is a set of knowledge concepts and $\mathcal{C}$ denotes text associated with or facts between the concepts.
One example of the knowledge base is Wikipedia, where concepts are entities proposed by users and text information of a concept corresponds to the article associated with the entity.
In general, our problem setting is applicable to any specific $\mathcal{C}$ as long as we can learn knowledge concept \textit{embeddings} from $\mathcal{C}$.
 Social text posted by users is denoted as $\mathcal{D}.$
Given a user $u \in \mathcal{V}^r$, $d_u \in \mathcal{D}$ denotes a document of all text posted by $u$. Each user $u$ has only one document $d_u$.
 
The output of the problem is a social knowledge graph $\mathcal{G} = (\mathcal{V}^r, \mathcal{V}^k, \mathcal{P})$. More specifically, given a user $u \in \mathcal{V}^r$, $\mathcal{P}_u$ is a ranked list of top-$k$ knowledge concepts in $\mathcal{V}^k$, where the order indicates the relatedness to user $u$. For example, in an academic social network, the algorithm outputs the top-$k$ research interests of each researcher $u$ as a ranked list $\mathcal{P}_u$.

There are two modalities in this problem, social network users and knowledge concepts. Previous problem settings usually consider only one of the two modalities. For example, social tag prediction \cite{heymann2008social} aims to assign tags to social network users without linking tags to knowledge bases. On the contrary, entity recognition in social context \cite{Tang:15WWW} extracts entities from social text without directly linking entities to users. In this sense, the problem of learning social knowledge graphs is technically challenging because we need to leverage information of both users and concepts.

 \section{Model Framework}
\label{sec:model}

We propose \textbf{GenVector}, a multi-modal Bayesian embedding model for learning social knowledge graphs. 
To jointly model multiple modalities, GenVector learns a shared latent topic space to generate network-based user embeddings and text-based concept embeddings in two different embedding spaces.

GenVector takes pretrained knowledge concept embeddings and user embeddings as input, where the pretrained embeddings encode information from the knowledge base $\mathcal{G}^k$ and the social network $\mathcal{G}^r$. In other words, embeddings are given as observed variables in our model. We use the Skip-gram model \cite{mikolov2013distributed} to learn knowledge concept embeddings, and use DeepWalk \cite{perozzi2014deepwalk} to learn network-based user embeddings.

\subsection{Generative Process}

\begin{figure}[tb]
	\centering
	\includegraphics[width=0.47\textwidth]{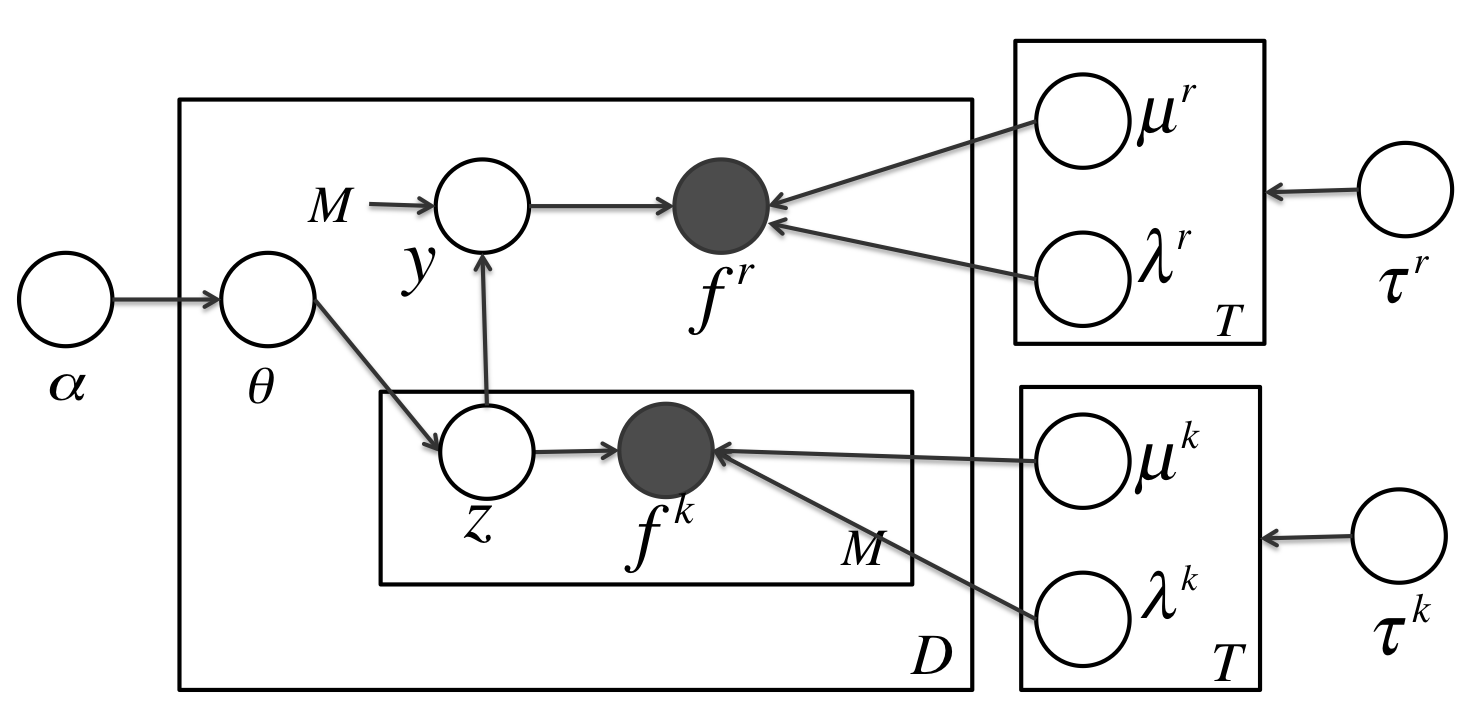}
	\caption{\label{fig:model}GenVector: a multi-modal Bayesian embedding model. Each document $d_u$ contains  all text posted by a social network user $u$ (each user has only one document). Embeddings are observed variables (dark circles).}
			
\end{figure}

GenVector is a generative model, and the generative process is illustrated in Figure~\ref{fig:model}, where we plot the graphical representation with $D$ documents and $T$ topics. Each document $d_u$ contains a user $u$ with all text posted by $u$.

The generative process is as follows:

\begin{enumerate}
\item For each topic $t$, and for each dimension
\begin{enumerate}
    \item Draw $\mu^r_t, \lambda^r_t$ from $\mbox{NormalGamma}(\tau^r)$
    \item Draw $\mu^k_t, \lambda^k_t$ from $\mbox{NormalGamma}(\tau^k)$
\end{enumerate}
\item For each user $u$
\begin{enumerate}
    \item Draw a multinomial distribution $\theta$ from $\mbox{Dir}(\alpha)$
    \item For each knowledge concept $w$ in $d_u$
    \begin{enumerate}
        \item Draw a topic $z$ from $\mbox{Multi}(\theta)$
        \item For each dimension of the embedding of $w$, draw $f^k$ from $\mathcal{N}(\mu_z^k, \lambda_z^k)$
    \end{enumerate}
    \item Draw a topic $y$ uniformly from all $z$'s in $d_u$
    \item For each dimension of the embedding of user $u$, draw $f^r$ from $\mathcal{N}(\mu_y^r, \lambda_y^r)$
\end{enumerate}
\end{enumerate}

\noindent where notations with superscript $^k$ denotes  parameters defined for knowledge concepts and notations with $^r$ for network users; $\tau$ is the hyperparameter of the normal Gamma distribution; $\mu$ and $\lambda$ are the mean and precision of the Gaussian distribution; $\alpha$ is the hyperparameter of the Dirichlet distribution; $\theta_u$ is the multinomial topic distribution of document $d_u$ (or user $u$); $z_{um}$ is the topic of the $m$-th knowledge concept in document $d_u$; $y_u$ is the topic of user $u$; $\mathcal{N}(\cdot)$ denotes the Gaussian distribution. Similarly, $f^k_{um}$ are knowledge concept embeddings, while $f^r_u$ are network-based user embeddings. We drop the subscripts or superscripts when there is no ambiguity.

Note that although it is possible to draw the embeddings from a multivariate Gaussian distribution, we draw each dimension from a univariate Gaussian distribution separately instead, because it is more computationally efficient and also practically performs well.
Although developed independently, our model extends Gaussian LDA \cite{das2015gaussian} to model multiple modalities.
Similar multi-modal techniques were also used in Corr-LDA \cite{blei2003modeling}. However, different from Corr-LDA, we generate continuous embeddings in two spaces and use normal Gamma distribution as the prior.

\subsection{Inference}
\label{sec:infer}

We employ collapsed Gibbs sampling \cite{griffiths2002gibbs} to do inference. The basic idea of collapsed Gibbs sampling is to integrate out the model parameters and then perform Gibbs sampling.
Due to space limitations, we directly give the conditional probabilities of the latent variables.

\begin{algorithm}[tb]
    \DontPrintSemicolon
    \small
    \KwIn{Training data $\mathcal{D}$, hyperparameters $\tau, \alpha$, initial embeddings $f^r, f^k$, burn-in iterations $t_b$, max iterations $t_m$, latent topic iterations $t_l$, parameter update period $t_p$}
    \KwOut{latent topics $z, y$, model parameters $\lambda, \mu, \theta$, updated embeddings $f^r, f^k$}

    \tcp{Initialization}
    random initialize $z, y$\;

    \tcp{Sampling}
    \For {$t \gets 1 \mbox{~to~} t_m$} {
        \For {$t' \gets 1 \mbox{~to~} t_l$} { \label{line:p1}
            \ForEach {latent topic $z$} {
                Draw $z$ according to Eq. (\ref{eq:z})\;
            }
            \ForEach {latent topic $y$} {
                Draw $y$ according to Eq. (\ref{eq:y})\;
            }
            \If {$t_p$ iterations since last read-out and $t > t_b$} {
                                Read out parameters according to Eq. (\ref{eq:param})\;
                Average all read-outs\; \label{line:p2}
            }
        } 

        \If {$t > t_b$} { \label{line:p3}
            Update the embeddings according to Eq. (\ref{eq:emb}) \label{line:up-emb}\;
        }
    }

    \Return{$z, y, \lambda^k, \lambda^r, \mu^k, \mu^r, \theta, f^r, f^k$}
    \normalsize
    \caption{Model Inference}
    \label{algo:infer}
\end{algorithm}

\small
\begin{equation}
p(y_u = t | y_{- u}, z, f^r, f^k) \propto (n_u^t + l) \prod_{e = 1}^{E^r} G'(f^r, y, t, e, \tau^r, u)
\label{eq:y}
\end{equation}
\begin{equation}
	\besp{
 &p(z_{um} = t | z_{- um}, y, f^r, f^k)   \\
&\propto (n_u^{y_u} + l) (n_u^t + \alpha_t) \prod_{e = 1}^{E^k} G'(f^k, z, t, e, \tau^k, um) 
}
\label{eq:z}
\end{equation}
\normalsize
where the subscript $_{-u}$ means ruling out dimension $u$ of a vector; $n_u^t$ is the number of knowledge concepts assigned to topic $t$ in document $d_u$; $E^r$ and $E^k$ are the dimensions of user embeddings and knowledge concept embeddings; $l$ is the smoothing parameter of Laplace smoothing \cite{manning2008introduction}.

The function $G'(\cdot)$ is given as follows
\[
G'(f, y, t, e, \tau, u) = \frac{\Gamma(\alpha_n)}{\Gamma(\alpha_{n'})} \frac{\beta_{n'}^{\alpha_{n'}}}{\beta_n^{\alpha_n}} \left(\frac{\kappa_{n'}}{\kappa_n}\right)^{\frac{1}{2}} \frac{(2\pi)^{- n / 2}}{(2\pi)^{- n' / 2}}
\]
with
\[
\alpha_n = \alpha_0 + n/ 2, ~~\kappa_n = \kappa_0 + n, ~~\mu_n = \frac{\kappa_0 \mu_0 + n \bar{x}}{\kappa_0 + n},
\]
\[
\beta_n = \beta_0 + \frac{1}{2}\sum_{i = 1}^{n_t} (x_i - \bar{x})^2 + \frac{\kappa_0 n (\bar{x} - \mu_0)^2}{2(\kappa_0 + n)}~~~~~~~~~~~
\]
where $\tau = \{\alpha_0, \beta_0, \kappa_0, \mu_0\}$ are the hyperparameters of the normal Gamma distribution; $n$ is the number of $i$'s with $y_i = t$; $x$ is a vector of concatenating the $e$-th dimension of $f_i$'s with $y_i = t$; $n' = n - 1$ if $y_u = t$, otherwise $n' = n$; $\bar{x}$ is the mean of all dimensions of $x$.

By taking the expectation of the posterior probabilities, we update the model parameters by
\[
\theta_u^t = \frac{n_u^t + \alpha_t}{\sum_{t' = 1}^T (n_u^{t'} + \alpha_{t'})}, ~~~~\mu_t = \frac{\kappa_0 \mu_0 + n \bar{x}}{\kappa_0 + n},
\]
\begin{equation}
\lambda_t = \frac{\alpha_0 + n / 2}{\beta_0 + \frac{1}{2} \sum_i (x_i - \bar{x})^2 + \frac{\kappa_0 n (\bar{x} - \mu_0)^2}{2(\kappa_0 + n)}}
\label{eq:param}
\end{equation}

In Gaussian LDA \cite{das2015gaussian}, the word embeddings were kept fixed during inference. Unlike their approach, in our model, we update the embeddings during inference to adapt to different specific problems.
Let $W$ be the number of knowledge concepts. We write the log likelihood of the data given the model parameters as
\beq{\small
	\besp{
L = \sum_{u = 1}^D \sum_{t = 1}^T \sum_{e = 1}^{E^r} & (- \frac{\lambda_{te}^r}{2}) (f_{ue}^r - \mu_{te}^r)^2 \\
&+ \sum_{w = 1}^W \sum_{t = 1}^T n_w^t \sum_{e = 1}^{E^k} (- \frac{\lambda_{te}^k}{2})(f^k_{we} - \mu_{te}^k)^2 
}
}
\normalsize

We employ gradient ascent to maximize the log likelihood by updating the embeddings $f^r$ and $f^k$. The gradients are computed as
\begin{equation}\small
\frac{\partial L}{\partial f_{ue}^r} = \sum_{t = 1}^T - \lambda_{te}^r (f_{ue}^r - \mu_{te}^r), ~\frac{\partial L}{\partial f_{we}^k} = \sum_{t = 1}^T n_w^t (- \lambda_{te}^k) (f_{we}^k - \mu_{te}^k)
\label{eq:emb}
\end{equation}\normalsize

The inference procedure is summarized in Algorithm~\ref{algo:infer}. Following \cite{heinrich2005parameter}, we set a burn-in period for $t_b$ iterations, during which we do not update embeddings or read out parameters. Similar to \cite{bezdek2003convergence}, we employ alternating optimization for inference. More specifically, we first fix the embeddings to sample the topics and infer the model parameters (Cf. Line~\ref{line:p1} -~\ref{line:p2}, Algorithm~\ref{algo:infer}). After a number of iterations, we fix the topics and parameters, and use gradient ascent to update the embeddings (Cf. Line~\ref{line:p3} -~\ref{line:up-emb}, Algorithm~\ref{algo:infer}). We repeat the procedure for a given number of iterations.

\subsection{Prediction}

Given a user $u$ and a knowledge concept $w$, let $g_{uw}$ denote whether $y_u$ is drawn from $z_w$. Conditioned on $u$ and the model parameters, we compute the joint probability of $g_{uw} = 1$ and generating the embedding $f^k_w$,
\begin{eqnarray}
&& p(g_{uw} = 1, f^k_w | f^r_u) \nonumber \\
&\propto& \sum_{t = 1}^T ( p(z_w = t) p(g_{uw} = 1) p (f^r_d | y_u = t) p(f^k_w | z_w = t) \nonumber \\
&\propto& \sum_{t = 1}^T \theta_u^t (n_u^w + l) \mathcal{N}(f^r_u | \lambda_t^r, \mu_t^r) \mathcal{N}(f^k_w | \lambda_t^k, \mu_t^k) \label{eq:pred}
\end{eqnarray}

For each user $u$, we rank the interacted knowledge concepts $w$ according to Eq. (\ref{eq:pred}) to obtain $\mathcal{P}_u$. In this way, we construct a social knowledge graph via learning the multi-modal Bayesian embedding model. Note that although it is possible to rank the knowledge concepts by deriving $p(f^k_w | f^r_u)$, our preliminary experiments show that using Eq. (\ref{eq:pred}) gives better results.

 \section{Experiments}

In this section, we perform a series of experiments to evaluate the proposed methods. We compare our models with state-of-the-art models on three datasets, and also design an online test on our system to demonstrate the effectiveness of our method.

\subsection{Data and Evaluation}
\label{sec:data}

We deploy our algorithm and run the experiments on AMiner\footnote{https://aminer.org/}, an online academic search system \cite{tang2008arnetminer}.
	The academic social network $\mathcal{G}^r$ is constructed by viewing each researcher as a user, and undirected edges represent co-authorships between researchers. There may be multiple edges between a pair of researchers if they collaborate multiple times. We use the publicly available English Wikipedia as the knowledge base $\mathcal{G}^k$. Each ``category'' or ``page'' in Wikipedia is viewed as a knowledge concept.
	We use the full-text Wikipedia corpus\footnote{https://dumps.wikimedia.org/enwiki/latest/} as the text information $\mathcal{C}$ to learn the knowledge concept embeddings. Social text $\mathcal{D}$ is derived from publications, where document $d_u$ for researcher $u$ contains all publications authored by $u$. If a publication has multiple authors, the publication is repeated for each researcher in their corresponding documents.
The basic statistics are shown in Table~\ref{tab:stat}.
We compare the following methods.

\begin{table}[tb]
\centering
\caption{Data Statistics}
\begin{tabular}{l | r}
\hline
\# Social network users & 38,049,189 \\
\# Publications & 74,050,920 \\
\# Knowledge concepts & 35,415,011 \\
Corpus size in bytes & 20,552,544,886 \\
\hline
\end{tabular}
\label{tab:stat}
\end{table}

\textbf{GenVector} is our model proposed in Section~\ref{sec:model}. We empirically set $\mu_0 = 0$, $\kappa_0 = 1\mbox{\sc{e}-}5$, $\beta_0 = 1$, $\alpha_0 = 1\mbox{\sc{e}}3$, $T = 200$, $\alpha = 0.25$.

\textbf{GenVector-E} is a variation of GenVector without updating the embeddings in Line~\ref{line:up-emb} of Algorithm~\ref{algo:infer}. We compare GenVector-E with GenVector to evaluate the benefit of embedding update.

\textbf{Sys-Base} is the original algorithm adopted by our system. Sys-Base first extracts key terms using a state-of-the-art NLP rule based extraction algorithm \cite{mundy2007microsoft}, and sorts the key terms by frequency.

\textbf{CountKG} extracts knowledge concepts from social text $\mathcal{D}$ by referring to the knowledge concept set $\mathcal{V}^k$, and ranks the concepts by appearance frequency.

\textbf{Author-Topic} learns an author-topic model \cite{rosen2004author} and ranks the knowledge concepts by
$
\sum_{t = 1}^T p(w | t) p (t | u)
$, where $t, u, w$ denote topic, user and knowledge concept respectively. We set $T = 200$, $\alpha = 0.25$.

\textbf{NTN} is a neural tensor network \cite{socher2013reasoning} that takes $f^k_w$ and $f^r_u$ as the input vector, and outputs the probability of $u$ matching $w$.
We perform cross validation to set the weighting factor $\lambda = 1\mbox{\sc{e}-}2$ and the slice size $k = 4$.

It is difficult in practice to directly evaluate the results of social knowledge graphs. Instead, we consider two strategies---offline evaluation on three data mining tasks and an online A/B test with live users.

\subsection{Offline Evaluation}

We collect three datasets for evaluation.
We first learn a social knowledge graph based on the data described in Section \ref{sec:data} (we discard the long-tailed users that do not appear in our evaluation datasets).
For each researcher $u$, we treat $\mathcal{P}_u$ as the research interests of the researcher. Then we use the collected datasets in the following sections to evaluate the precision of the research interests.

\subsubsection{Homepage Matching}
\label{sec:homepage}

We crawl $62,127$ researcher homepages from the web. After filtering out those pages that are not informative enough (\# knowledge concepts $< 5$), we obtain $1,874$ homepages. We manually identify the research interests that are explicitly specified by the researcher on the homepage, and treat those research interests as ground truth. We then evaluate different methods based on the ground truth and report the precision of the top 5 knowledge concepts. The performances are listed in Table~\ref{tab:homepage}.
GenVector  outperforms Sys-Base, Author-topic, and NTN by $5.8\%$, $4.9\%$, and $18.5\%$ respectively.

\begin{table}[tb]
\centering
\caption{Precison@5 of Homepage Matching}
\begin{tabular}{l l}
\hline
Method & Precision@5 \\
\hline
\textbf{GenVector} & \textbf{78.1003\%} \\
GenVector-E & 77.8548\% \\
Sys-Base & 73.8189\% \\
Author-Topic & 74.4397\% \\
NTN & 65.8911\% \\
CountKG & 54.4823\% \\
\hline
\end{tabular}
\label{tab:homepage}
\end{table}

By comparing NTN with GenVector, we show that taking the learned embeddings as input without exploiting the latent topic structure cannot result in good performance, although NTN is among the most expressive models given plain vectors as input \cite{socher2013reasoning}. NTN does not perform well because it has no prior knowledge about the underlying structure of data, and it is thus difficult to learn a mapping from embeddings to a matching probability.
GenVector performs better than Author-Topic, which indicates that incorporating knowledge concept embeddings and user embeddings can boost the performance. In this sense, GenVector successfully leverages both network structure (by learning the user embeddings) and large-scale unlabeled corpus (by learning the knowledge concept embeddings).

GenVector also significantly outperforms Sys-Base and CountKG. Sys-Base and CountKG compute the importances of the knowledge concepts by term frequency. For this reason, the extracted knowledge concepts are not necessarily semantically important. Sys-Base is better than CountKG because Sys-Base uses the key term extraction algorithm \cite{mundy2007microsoft} to filter out frequent but unimportant knowledge concepts.

The difference between GenVector-E and GenVector indicates that updating the embeddings can further improve the performance of the proposed model. This is because updating the embeddings to fit the data in specific problems, is better than using general embeddings learned from unlabeled data.

\subsubsection{LinkedIn Profile Matching}

We design another experiment to evaluate the methods based on the LinkedIn profiles of researchers. We employ the network linking algorithm COSNET \cite{zhang2015cosnet} to link the academic social network on our system to the LinkedIn network. More specifically, given a researcher on our system, COSNET finds the according profile on LinkedIn, if any.

We first select the connected pairs with highest probabilities given by COSNET, and then manually select the correct ones. We use the selected pairs as ground truth, e.g., A on our system and B on LinkedIn are exactly the same researcher in the physical world.

Some LinkedIn profiles of researchers have a field named ``skills'', which contains a list of expertise. Once a researcher accept endorsements on specific expertise from their friends, the expertise is appended to the list of ``skills''. After filtering out researchers with less than five ``skills'', we obtain a dataset of $113$ researchers. We use the list of ``skills'' as the ground truth of research interests. We report the precision of top 5 research interests in Table~\ref{tab:lk}. Since some of the ``skills'' are not necessarily research interests (e.g. Python), we focus on precision and do not consider recall-based evaluation metrics.

\begin{table}[tb]
\centering
\caption{Precision@5 of LinkedIn Profile Matching}
\begin{tabular}{l l}
\hline
Method & Precision@5 \\
\hline
\textbf{GenVector} & \textbf{50.4424\%} \\
GenVector-E & 49.9145\% \\
Author-Topic & 47.6106\% \\
NTN & 42.0512\% \\
CountKG & 46.8376\% \\
\hline
\end{tabular}
\label{tab:lk}
\end{table}

We can observe from Table~\ref{tab:lk} that GenVector gives the best performance. GenVector outperforms CountKG, Author-Topic and NTN by $7.7\%$, $5.9\%$, and $20.0\%$ respectively (sign test over samples $p < 0.05$). Updating the embeddings improves the performance by $1.1\%$.

\subsubsection{Intruder Detection}

In this experiment, we employ human efforts to judge the quality of the social knowledge graph. Since annotating research interests is somewhat subjective, we label the research interests that are clearly not relevant \cite{liu2009unsupervised}, also known as intruder detection \cite{chang2009reading}. In other words, instead of identifying the research interests of a researcher, we label what are definitely NOT the research interests of a researcher, e.g., ``challenging problem'' and ``training set''.

We randomly pick $100$ high cited researchers on our system. For each researcher, we run different algorithms to output a ranked list of knowledge concepts. We combine the top 5 knowledge concepts of each algorithm and perform a random shuffle. The labeler then labels clearly irrelevant research interests in the given list of knowledge concepts. We report the error rate of each method in Table~\ref{tab:human}.

\begin{table}[tb]
\centering
\caption{Error Rate of Irrelevant Cases}
\begin{tabular}{l r}
\hline
Method & Error Rate \\
\hline
\textbf{GenVector} & \textbf{1.2\%} \\
Sys-Base & 18.8\% \\
Author-Topic & 1.6\% \\
NTN & 7.2\% \\
\hline
\end{tabular}
\label{tab:human}
\end{table}

According to Table~\ref{tab:human}, GenVector produces less irrelevant knowledge concepts than other methods. It is because GenVector leverages large-scale unlabeled corpus to encode the semantic information into the embeddings, and therefore is able to link researchers to major research interests.

\subsection{Online Test}

To further test the performance of our algorithm, we deploy GenVector on our online system with the full dataset described in Section \ref{sec:data}. We leverage collective intelligence by asking the users to select what they think are the research interests of the given researcher.

\begin{figure}[tb]
\centering
\includegraphics[width=0.4\textwidth]{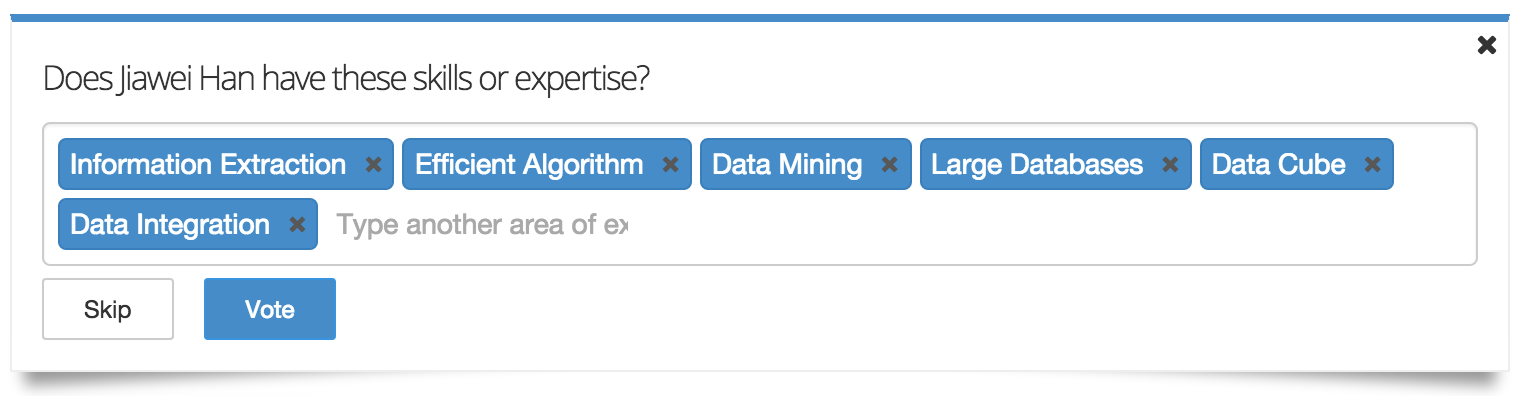}
\caption{Questionnaire: Leveraging Collective Intelligence for Evaluation}
\label{fig:ques}
\end{figure}

Since Sys-Base is the original algorithm adopted by our system, we perform an online test by comparing GenVector with Sys-Base to evaluate the performance gain. For each researcher, we first compute the top 10 research interests provided by the two algorithms. Then we randomly select 3 research interests from each algorithm, and merge the selected research interests in a random order. When a user visits the profile page of a researcher, a questionnaire is displayed on top of the profile. A sample is shown in Figure~\ref{fig:ques}. Users can vote for research interests that they think are relevant to the given researcher.

\begin{table}[tb]
\centering
\caption{Error Rate of Online Test}
\begin{tabular}{l r}
\hline
Method & Error Rate \\
\hline
\textbf{GenVector} & \textbf{3.33\%} \\
Sys-Base & 10.00\% \\
\hline
\end{tabular}
\label{tab:online}
\end{table}

We collect $110$ questionnaires in total, and use them as ground truth to evaluate the algorithms. The error rates of different algorithms are shown in Table~\ref{tab:online}. We can observe that GenVector decreases the error rate by $67\%$. Moreover, the error rate of GenVector is lower than or equal to that of Sys-Base for $95.45\%$ of the collected questionnaires (sign test over samples $p \ll 0.01$).

\subsection{Run Time and Convergence}

\begin{figure}[tb]
\centering
\includegraphics[width=0.34\textwidth]{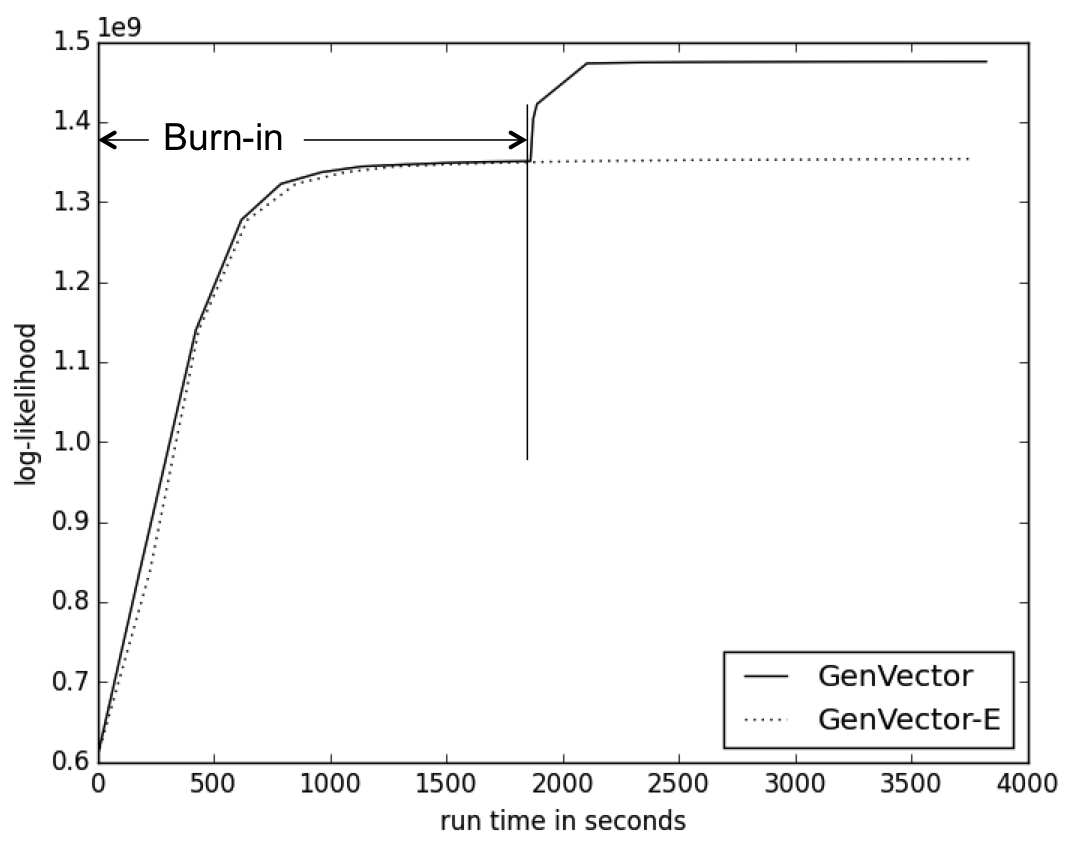}
\caption{Run Time and Convergence: log-likelihood v.s. run time in seconds.}
\label{fig:time}
\end{figure}

Figure~\ref{fig:time} plots the run time and convergence of GenVector and GenVector-E. During the burn-in period, GenVector and GenVector-E perform identically because GenVector does not update the embeddings during the period. After the burn-in period, the likelihood of GenVector continues to increase while that of GenVector-E remains stable, which indicates that by updating the embeddings, GenVector can better fit the data, which leads to better performance shown in previous sections. The experiments were run on Intel(R) Xeon(R) CPU E5-4650 0 @ 2.70GHz with 64 threads.

\subsection{Case Study}

Table~\ref{tab:topics} shows the researchers and knowledge concepts within each topic, output by GenVector and Author-Topic, where each column corresponds to a topic. As can be seen from Table~\ref{tab:topics}, Author-Topic identifies several irrelevant concepts (judged by human) such as ``integrated circuits'' in topic \#1, ``food intake'' in topic \#2, and ``in vitro'' in topic \#3, while GenVector does not have this problem.

 \section{Related Work}

Variants of topic models \cite{hofmann1999probabilistic,blei2003latent} represent each word as a vector of topic-specific probabilities. Although Corr-LDA \cite{blei2003modeling} and the author-topic model \cite{rosen2004author} can be used for multi-modal modeling, the topic models use discrete representation for observed variables, and are not able to exploit the continuous semantics of words and authors.

Learning embeddings \cite{mikolov2013distributed,levy2014neural,perozzi2014deepwalk} is effective at modeling continuous semantics with large-scale unlabeled data, e.g., knowledge bases and network structure. Neural tensor networks \cite{socher2013reasoning} are expressive models for mapping the embeddings to the prediction targets. However, GenVector can better model multi-modal data by basing the embeddings on a generative process from latent topics.

Recently a few research works \cite{das2015gaussian,wan2012hybrid} propose hybrid models to combine the advantages of topic models and embeddings. Gaussian embedding models \cite{vilnis2014word} learn word representation via a Gaussian generative process to encode hierarchical structure of words. However, these models are proposed to address other issues in semantic modeling, and cannot be directly used for multi-modal data.

Learning social knowledge graphs is also related to keyword extraction. Different from conventional keyword extraction methods \cite{liu2009unsupervised,mundy2007microsoft,matsuo2004keyword,rao2013entity}, our method is based on topic models and embedding learning. \begin{table}[tb]
\tiny
\centering
\caption{Knowledge Concepts and Researchers of Given Topics. $*$ marks relatively irrelevant concepts.}
\begin{tabular}{l|l|l}
\hline
Topic \#1 & Topic \#2 & Topic \#3 \\
\hline
\multicolumn{3}{c}{GenVector} \\
\hline
query expansion&image processing&hepatocellular carcinoma\\
concept mining&face recognition&gastric cancer\\
language modeling&feature extraction&acute lymphoblastic leukemia\\
information extraction&computer vision&renal cell carcinoma\\
knowledge extraction&image segmentation&glioblastoma multiforme\\
entity linking&image analysis&acute myeloid leukemia\\
language models&feature detection&peripheral blood\\
named entity recognition&digital image processing&malignant melanoma\\
document clustering&machine learning algorithms&hepatitis c virus\\
latent semantic indexing&machine vision&squamous cell carcinoma\\
\hline
Thorsten Joachims&Anil K. Jain&Keizo Sugimachi\\
Jian Pei&Thomas S. Huang&Setsuo Hirohashi\\
Christopher D. Manning&Peter N. Belhumeur&Masatoshi Makuuchi\\
Raymond J. Mooney&Azriel Rosenfeld&Morito Monden\\
Charu C. Aggarwal&Josef Kittler&Yoshio Yamaoka\\
William W. Cohen&Shuicheng Yan&Kunio Okuda\\
Eugene Charniak&David Zhang&Yasuni Nakanuma\\
Kamal Nigam&Xiaoou Tang&Kendo Kiyosawa\\
Susan T. Dumais&Roberto Cipolla&Masazumi Tsuneyoshi\\
T. K. Landauer&David A. Forsyth&Satoru Todo\\
\hline
\multicolumn{3}{c}{Author-Topic} \\
\hline
speech recognition&face recognition&hepatocellular carcinoma\\
natural language&$^*$food intake&kidney transplantation\\
$^*$integrated circuits&face detection&cell line\\
document retrieval&image recognition&differential diagnosis\\
language models&$^*$atmospheric chemistry&liver tumors\\
language model&feature extraction&cell lines\\
$^*$microphone array&statistical learning&squamous cell carcinoma\\
computational linguistics&discriminant analysis&$^*$in vitro\\
$^*$semidefinite programming&object tracking&kidney transplant\\
active learning&$^*$human factors&lymph nodes\\
\hline
James F. Allen&Anil K. Jain&Keizo Sugimachi\\
Christopher D. Manning&Kevin W. Bowyer&Giuseppe Remuzzi\\
Eugene Charniak&David Zhang&Setsuo Hirohashi\\
T. K. Landauer&Xiaoou Tang&H. Fujii\\
Andrew B. Kahng&Ming-Hsuan Yang&Paul I. Terasaki\\
A. Sangiovanni-Vincentelli&S. Shankar Sastry&M. Watanabe\\
Partha Niyogi&P. J. Crutzen&Robert A. Wolfe\\
Lillian Lee&Stan Z. Li&David E. R. Sutherland\\
Daniel Jurafsky&Keith W. Ross&G. Chen\\
Zhi-Quan Luo&Jingyu Yang&Arthur J. Matas\\
\hline
\end{tabular}
\label{tab:topics}
\end{table} \section{Conclusion}

In this paper, we study the problem of learning social knowledge graphs. We propose GenVector, a multi-modal Bayesian embedding model, to jointly incorporate the advantages of topic models and embeddings. GenVector models the network embeddings and knowledge concept embeddings in a shared topic space. We present an effective learning algorithm that alternates between topic sampling and embedding update. Experiments show that GenVector outperforms state-of-the-art methods, including topic models, embedding-based models and keyword extraction based methods. We deploy the algorithm on a large-scale social network and decrease the error rate by 67\% in an online test.

 
 \small
\vpara{Acknowledgements.}Zhilin Yang and Jie Tang are supported by 
973 (2014CB340506) and 863 (
2015AA124102).
 
 \normalsize

\bibliographystyle{named}
\bibliography{main}

\end{document}